\documentclass[journal]{IEEEtai}

\usepackage[colorlinks,urlcolor=blue,linkcolor=blue,citecolor=blue]{hyperref}

\usepackage{color,array}
\usepackage{graphicx}

\usepackage{graphicx}
\usepackage{listings}
\usepackage{subfigure}
\usepackage{amsmath}
\usepackage{rotating}
\usepackage{multirow}
\usepackage{lscape}
\usepackage{verbatim}
\usepackage{hhline}
\usepackage{textcomp,booktabs}
\usepackage{xcolor,colortbl}
\usepackage{paralist}
\usepackage{textcomp}
\usepackage{wrapfig}
\usepackage{amsthm}
\usepackage{url}
\usepackage{breakurl}
\usepackage{caption}
\usepackage[ruled,vlined,linesnumbered]{algorithm2e}

\usepackage{colortbl}
\usepackage{array}

\newcolumntype{P}[1]{>{\centering\arraybackslash}p{#1}}
\newcolumntype{M}[1]{>{\centering\arraybackslash}m{#1}}

\begin{document}

\title{FedAVO: Improving Communication Efficiency in Federated Learning with African Vultures Optimizer 
}

\author{Md Zarif Hossain and Ahmed Imteaj\\School of Computing, Southern Illinois University Carbondale, IL 62901, USA
\thanks{Md Zarif Hossain is a graduate research assistant of Security, Privacy and intElligence for Edge Devices (SPEED) Lab and a PhD student at Southern Illinois University Carbondale, IL 62901 USA (e-mail: mdzarif.hossain@siu.edu).}
\thanks{Ahmed Imteaj is the director of Security, Privacy and intElligence for Edge Devices (SPEED) Lab and an Assistant Professor at Southern Illinois University Carbondale, IL 62901, USA (e-mail: imteaj@cs.siu.edu).}

\thanks{This paragraph will include the Associate Editor who handled your paper.}}

 \markboth{}
{Md Zarif Hossain \MakeLowercase{\textit{et al.}}: FedAVO: Improving Communication Efficiency in Federated Learning with African Vultures Optimizer}

\maketitle

\begin{abstract}
Federated Learning (FL) has recently experienced tremendous popularity due to its emphasis on user data privacy. However, the distributed computations of FL can result in constrained communication and drawn-out learning processes, necessitating the client-server communication cost optimization. The ratio of chosen clients and the quantity of local training passes are two hyperparameters that have a significant impact on the performance of FL. Due to different training preferences across various applications, it can be difficult for FL practitioners to manually select such hyperparameters. In this paper, we introduce FedAVO, a novel FL algorithm that enhances communication effectiveness by selecting the best hyperparameters leveraging the African Vulture Optimizer (AVO). Our research demonstrates that the communication costs associated with FL operations can be substantially reduced by adopting AVO for FL hyperparameter adjustment. Through extensive evaluations of FedAVO on benchmark datasets, 
we identify the optimal hyperparameters that are appropriately fitted for the benchmark datasets, eventually increasing global model accuracy by 6\% in comparison to the state-of-the-art FL algorithms (such as FedAvg, FedProx, FedPSO). The code, data, and experiments have been made publicly available on our GitHub repository\footnote{\url{https://github.com/speedlab-git/FedAVO}}.
\end{abstract}

\begin{IEEEImpStatement}
This paper introduces a novel and impactful approach to enhancing communication efficiency in FL. Our proposed FedAVO method addresses the challenge of communication overhead in FL, paving the way for significant advancements in distributed machine learning. This novel solution reduces communication rounds, thereby optimizing resource utilization, and also tailors a creative adaptation of nature-inspired optimization techniques for practical computing applications. The potential impact of FedAVO extends to various domains, including privacy-preserving FL, edge computing, and applications requiring real-time model updates such as such as connected autonomous vehicles. This research contributes not only to the technical landscape of FL but also holds promise for improving the scalability, speed, and sustainability of federated learning methodologies.

\end{IEEEImpStatement}


\section{Introduction}
\label{sec:1}
\IEEEPARstart{T}{he} proliferation of diverse consumer electronics, including IoT devices, home appliances, smartphones, connected autonomous vehicles (CAVs), drones, Virtual Reality (VR), and Augmented Reality (AR) devices, has witnessed substantial growth in recent years. These devices capture and store various forms of data such as images, audio, and text.Image, audio, and text are just a few of the different forms of data that such devices acquire and store.  
Every day brings a slight increase in the number of IoT-related applications, their users, and the amount of data they produce. This phenomenon is pioneering the scope of data-driven decision-making. Due to the heterogeneous nature of these collected data, training machine learning models comes with the following challenges: \textit{widely distributed}: data points are stored in a large number of clients, which can be significantly higher than the average number of training samples retained on a given client; \textit{non-IID Data}: each client's data may not accurately represent the entire distribution \cite{mcmahan2017communication} of a dataset; \textit{unbalanced Data}: clients may hold different magnitudes of data points; \textit{communication constraints}: Mobile or IoT devices frequently may experience poor, expensive, or no internet connectivity; \textit{communication cost}: data collection from edge devices is significantly high. Moreover, transferring the data to the central server for training requires additional network bandwidth.

The study FL has made substantial progress in addressing the aforementioned challenges. FL is a machine learning (ML) paradigm that trains ML models utilizing decentralized data. FL protects data privacy by preventing data transfer from local machines to a centralized server. Each edge devices train the model on its own data locally using its computation resources. Instead of sending local data to the central server, only the local updates are sent. By providing only the local updates, FL cuts down the bandwidth overhead and also the communication cost. For these distinctive qualities, FL has been adopted in a wide range of applications such as speech recognition \cite{paulik2021federated, tomashenko2022privacy}, surveillance system \cite{jiang2022federated, brik2020federated}, health care system \cite{xu2021federated, antunes2022federated}, Internet of Things (IoT)\cite{imteaj2021survey}, human stroke prevention \cite{ju2020privacy, joshi2022federated}, and cybersecurity \cite{khramtsova2020federated}. An FL paradigm includes local training, client-server communication, and model aggregation \cite{mcmahan2017communication, wang2020tackling}. Often communication overhead in FL comes from model broadcast from the server to all clients and vice versa. Furthermore, there is a feasibility risk in every communication round in terms of constrained network bandwidth, packet transmission loss, and privacy invasion. Apart from communication overhead and privacy invasion, training on heterogeneous setups opposes some more challenges. After the centralized server broadcasts its model, the clients train the model while considering some hyperparameters such as learning rate, batch dimension, and rounds per epoch. These diversely trained client models are difficult to aggregate since computational power and data attributes (complexity, ambiguity) differ vastly between each edge device. In an idealistic scenario of hundreds, even thousands of devices, the updated global model may never converge to a global optimum due to the heterogeneous behavior of the clients. Existing aggregation techniques (e.g., FedAvg \cite{mcmahan2017communication}, FedMa \cite{wang2020federated}) only focus on integrating weights of local models. To increase generalization and convergence rate, some novel approaches suggested model aggregation by using feature fusion \cite{yao2019towards} of global and local models or using multiple global models \cite{kopparapu2020fedcd}. 
\vspace{-0.1cm}

Along with the typical hyperparameters of model training, such as learning rates, optimizers, and mini-batch sizes, FL also has particular hyperparameters, such as local epochs and participant selection. The selection of these hyperparameters can drastically affect the performance of FL. The impact of hyperparameter tuning is much higher in Non-IID data distribution because data samples are radically different from each other across the clients. Although hyperparameter tuning optimization (HTO) has been widely explored in centralized machine learning, several aspects of HTO in FL settings yet need to be studied. In centralized machine learning, the model often trains on the entire dataset, which is often not viable in FL settings. In addition to that, models train on a wide range of hyperparameter configurations, which is extremely expensive in terms of communication and training time in FL settings because each FL cycle consists of multiple phases and communication rounds, and the model must complete each communication round in order to evaluate or change any hyperparameters. Finally, the same hyperparameter configuration functions less well as centralized machine learning due to the heterogeneous behavior of the clients and their data in FL settings. A few strategies for FL-HTO have recently been put forth; however, they concentrate on handling HTO using personalization techniques and neural networks, and they often lack in optimizing the communication cost and convergence rate. In this study, we optimize automated hyperparameter tuning with the help of a meta-heuristic algorithm. We propose FedAVO, which can configure and tune hyperparameters locally concerning the data distribution and their quality. To summarize, our proposed work outlines three novel contributions, which are listed as follows:\\
$\bullet$ To the best of our knowledge, FedAVO is the first FL hyperparameter tuning algorithm that reveals the potential of AVO as an optimizer for automated hyperparameter tuning during FL model training, specifically tailored for consumer electronics functioning as edge devices.\\
$\bullet$ FedAVO can be employed to tune the hyperparameters automatically for any state-of-the-art FL algorithm (such as FedAvg \cite{mcmahan2017communication}, FedProx \cite{li2020federated}, and FedNova \cite{wang2020tackling}). Moreover, FedAVO can automatically configure and tune the hyperparameters locally on each communication rounds.\\
$\bullet$ We conduct an in-depth evaluation of FedAVO, comparing its performance with the state-of-the-art FL algorithms. Our findings reveal a notable decrease in communication overhead and an increase in convergence rate. Compared to FedAvg, FedAVO reduces the requirement of communication rounds to reach convergence by 80 on average. In addition, FedAVO converges much faster and achieves better accuracy than the state-of-the-art FL algorithms. For instance, FedAVO achieves 99.47\% accuracy with MNIST Non-IID distribution and 72.64\% accuracy with CIFAR-10 Non-IID distribution.  Moreover, for Fashion MNIST and LISA datasets, FedAVO achieves consecutive global accuracies of 85.28\% and 92.88\%, respectively.
The rest of the paper is organized as follows:
Section \ref{sec:2} depicts the fundamental concepts of FL and the key elements of our proposed method: AVO algorithm. Related works are reviewed in Section \ref{sec:3}, while Section \ref{sec:method} explains the algorithm and operating procedure of FedAVO. Section \ref{sec:4} presents the experimental analysis and highlights the significant findings. Finally, the study is concluded with a discussion of future directions to enhance our work in section \ref{sec:5}.

\section{Background Study}
\label{sec:2}
African Vulture Optimization Algorithm (AVOA) is a nature inspired population based metaheuristic algorithm proposed in \cite{abdollahzadeh2021african}. The algorithm was derived from the behavior and characteristics of African vultures, leveraging their adaptive and efficient foraging strategies observed in the natural world. AVOA offers a promising method for addressing complex optimization challenges across wide-range of domains.
We presented the steps of the AVOA in below:

\subsubsection{Population Classification}
\begin{figure}[htb!]
\setlength{\belowcaptionskip}{-10pt}
  \centering
  \includegraphics[width=0.75\linewidth]{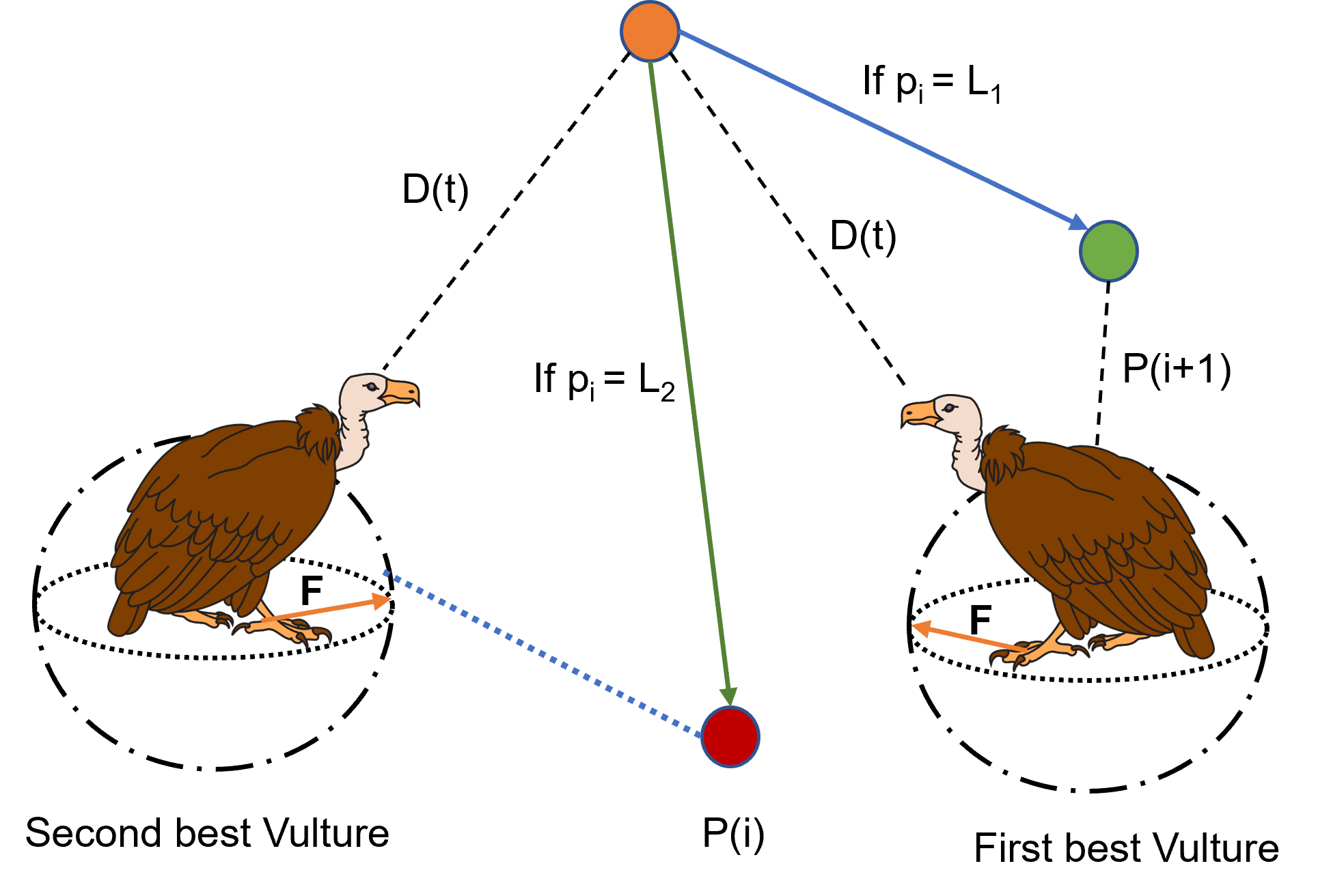}
    \caption{{\color{black}Illustration of African Vulture Optimization technique.}}
    \label{fig:2}  
\end{figure}

African vultures are categorized into three groups based on their living habits, as discussed in the cited references \cite{yakout2022comparison, bagal2021sofc}. The first group comprises vultures that represent optimal solutions when assessing their fitness with respect to feasible solutions. The second group includes vultures representing the second-best feasible solutions, while the remaining vultures are placed in the third group. Figure \ref{fig:2} illustrates the AVO architecture, with the right vulture symbolizing the best vulture and the left one representing the second-best. The fitness of all vulture populations is evaluated after establishing the initial population, and they are categorized based on their fitness values using Equation \ref{eq:1}.
\begin{equation}
R_i^t= \begin{cases}\text { Best vulture }{ }_1^t, & \text { if } P_i^t=L_1 \\ \text { Best vulture }{ }_2^t, & \text { if } P_i^t=L_2\end{cases}
\label{eq:1}
\end{equation}
Here, Best $vulture_1$ represents the best optimal solution for the population and Best $vulture_2$ depicts the suboptimal solution. 
$L_1$ and $L_2$ are both random values within the range of 0 to 1, and their sum equals to 1. 
With the help of the roulette-wheel technique, $p_i$ is calculated using equation \ref{eq:2}: 
\begin{equation}
p_i^t=\frac{F_i^t}{\sum_{i=1}^n F_i^t}
\label{eq:2}
\end{equation}
Here, the fitness value of vultures is represented by $F_i$ and n represents the total vultures in the first and second groups.
\subsubsection{Rate of Starvation in Vultures}
In FL with AVO, we can liken vultures to computational entities or clients in the learning process. Well-fed vultures represent entities with ample resources and energy, allowing them to explore the optimization landscape extensively and perform resource-intensive computations. They handle complex tasks in Federated Learning. Conversely, starving vultures symbolize resource-constrained or energy-depleted entities, lacking capacity for prolonged and resource-intensive computations, similar to hungry vultures unable to sustain extended flights. In FL, these entities may face limitations like restricted resources or communication bandwidth. Starvation rate among vultures significantly affects the optimization process's exploration and development phases. Changes in this rate directly impact collaboration, strategy adaptation, and computational allocation within the FL ecosystem. Ultimately, these dynamics shape the optimization process in FL with AVO
The vulture starvation rate $S_i$ is computed by the following equation \ref{eq:3}:
\begin{equation}
S_i=\left(2 \times \operatorname{rand}_i+1\right) \times h \times\left(1-\frac{\text { iteration }_i}{\text { maxiterations }}\right)+t
\label{eq:3}
\end{equation}
\vspace{-0.2cm}
\subsubsection{Exploration Stage}
African vultures take some time to locate other populations and food before making a long flight searching for it. They might face a hard time locating food since they spend a long time exploring their surroundings before flying farther distances in the quest of food. In AVOA, two distinct exploration strategies are discussed, and $P_1$ in the range of [0,1] is used to determine the strategy to be adopted. A random number $rand_{p1}$ ranging between 0 and 1 is utilized to select one of the strategies during the exploration phase. The random value gets compared to the value of $P_1$; if the random value is lesser than $P_1$,  equation \ref{eq:5} is utilized; otherwise, equation \ref{eq:6} is utilized to determine the strategy:
\begin{equation}
P_i^{t+1}=
R_i^t-D_i^t \times S \quad \text { if } P_1 \geq \operatorname{rand}_{p 1} 
\label{eq:5}
\end{equation}
\begin{equation}\small
 P_i^{t+1}=R_i^t-S+\operatorname{rand}_2 \times\left((u b-l b) \times \text {rand}_3+l b\right) \\
\text { if } P_1<\operatorname{rand}_{p 1} 
\label{eq:6}
\end{equation}



\subsubsection{Development Stage}
In this stage, the AVO starts the first stage of exploitation, if $|S|$ is between 0.5 and 1, the vultures enter the initial development stage and during this stage, they search for prey in two strategies. $P_2$ ranging between 0 and 1, determines whether the vultures compete for food or spiral. $round_{p_2}$, a random number between 0 and 1 gets generated at the first stage of this phase. If $round_{p_2}$ is greater than or equal to parameter $P_2$ the siege-fight strategy is applied slowly. Otherwise, the rotational flying technique is utilized. equation \ref{eq:8} and \ref{eq:extend} demonstrates this procedure:
\begin{equation}
P^{t+1}_i= D^t_i \times\left(S_i+\text { rand }_4\right)- R^t_i- P^t_i  \\\quad\quad\text { if } P_2 \geq \operatorname{rand}_{p_2} \\
\label{eq:8}
\end{equation}
\begin{equation}
P^{t+1}_i=R^t_i-\left(Q^t_1+Q^t_2\right) \\ \quad\quad\text { if } P_2<\operatorname{rand}_{p_2}
\label{eq:extend}
\end{equation}
where, $rand_4$ is a random number ranging between 0 and 1, and the term ($R^t_i- P^t_i$) is used for calculating the distance between the vulture and one of the two groups' best vultures:
\begin{equation}
 Q_1^t=R_i^t \times\left(\frac{\text { rand }_5 \times P_i^t}{2 \pi}\right) \times \cos \left(P_i^t\right)
 \label{eq:9}
\end{equation}

\begin{equation}
 Q_2^t=R_i^t \times\left(\frac{\text { rand }_6 \times P_i^t}{2 \pi}\right) \times \sin \left(P_i^t\right)
 \label{eq:10}
\end{equation}
\subsubsection{Final Stage (Second phase)}
The algorithm shifts to this stage when the $|S_i|$ value is less than 0.5. It indicates all vultures are full, but the best two types of vultures become hungry and weak after a prolonged flying session. At this moment, vultures attack their prey, and multiple vultures congregate at the same food source. Parameter $P_3$, ranging between 0 and 1, is used to determine whether the vulture shows aggregation behavior or aggressive behavior. When the vulture enters the second phase of the final stage, $rand_{p3}$ gets initialized with a value between 0 and 1. When $rand_{p3}$ is less than $P_3$, the vulture engages in aggressive behavior; Otherwise, engages in aggregating behavior. This equation governs their movement and behavior during this critical phase of the optimization process. The vultures position can be updated using following equations \ref{eq:11}, \ref{eq:12} and \ref{eq:13}:
\begin{equation}
P_i^{t+1} =\frac{A_1^t+A_2^t}{2}
\label{eq:11}
\end{equation}
\begin{equation}
    A_1^t =\text { Best Vulture }{ }_1^t-\frac{\text { Best Vulture }{ }_1^t \times P_i^t}{\text { Best Vulture }{ }_1^t-\left(P_i^t\right)^2} \times S \\
    \label{eq:12}
\end{equation}

\begin{equation}
    A_2^t  =\text { Best Vulture }{ }_2^t-\frac{\text { Best Vulture }{ }_2^t \times P_i^t}{\text { Best Vulture }_2^t-\left(P_i^t\right)^2} \times S \\
    \label{eq:13}
\end{equation}
Here, ${ Best Vulture }{ }_1^t$ and ${ Best Vulture }{ }_2^t$ are the optimal and suboptimal solutions and S is the starvation rate. 
\section{Related Works}
\label{sec:3}
The rising prevalence of smart and IoT devices in daily life has amplified the need for FL, primarily due to its capability to safeguard data privacy during the model training process. Consequently, there has been a surge in the exploration of efficient optimization techniques tailored to FL.
Researchers have directed their attention to various facets of FL, encompassing aspects such as communication efficiency \cite{basu2019qsparse,
laguel2021superquantile},  client selection \cite{imteaj2020fedar, xu2020client}, the intricate dynamics of statistical and system heterogeneity \cite{karimireddy2020mime,li2019fedmd} and cybersecurity \cite{shahid2023assessing}. One particular area of interest that aligns with our research focus revolves around the reduction of communication rounds in federated learning. Our approach to achieving this reduction centers on the optimization of hyperparameters.
These hyperparameters, pivotal components of a model, govern its ability to glean insights from a specific dataset.In order to achieve better performance, the hyperparameters need to be fine-tuned with different datasets and different models. 
The most straightforward automated approach for adjusting hyperparameters involves conducting random searches \cite{bergstra2012random} within the hyperparameter space to identify the best-performing set. However, this method was surpassed by Bayesian techniques \cite{bergstra2011algorithms}, which leverage the performance of previously chosen hyperparameters to inform the selection of new ones. Notably, these methods demand substantial computational resources, as the complete training process must be executed to assess hyperparameter fitness. To mitigate the computational burden, there is a shift toward employing partial training \cite{klein2017fast, li2017hyperband} with pre-selected hyperparameters. Nevertheless, even with this approach, running full or partial training to identify optimal hyperparameters within resource-constrained FL environments, where communication budgets are limited, remains prohibitively costly and cumbersome.
While several researchers have explored the optimization of machine learning (ML) models and their hyperparameters using evolutionary algorithms \cite{kim2019evolutionary}, including techniques like whale optimization \cite{aljarah2018optimizing} and genetic algorithms \cite{xiao2020efficient}, the same level of attention has not been given to FL. There have been only a few attempts to address FL HTO problems. For instance, the FLoRA framework \cite{zhou2021flora} introduces a novel HTO strategy where global hyperparameters are determined by selecting well-performing hyperparameters from local clients. FedEx \cite{khodak2021federated}, on the other hand, optimizes hyperparameter tuning by leveraging Neural Architecture Search (NAS) techniques with a weight-sharing architecture.
Nevertheless, several optimization approaches encounter challenges in certain FL scenarios by inadvertently introducing overhead during hyperparameter tuning. Our proposed HTO method strives to bridge the existing gaps, presenting an efficient approach to optimize hyperparameters within the FL framework.
\vspace{-0.2cm}

\section{Proposed Framework: FedAVO}
\label{sec:method}
\begin{figure*}[htb!]
  \centering
  \includegraphics[width=0.7\textwidth]{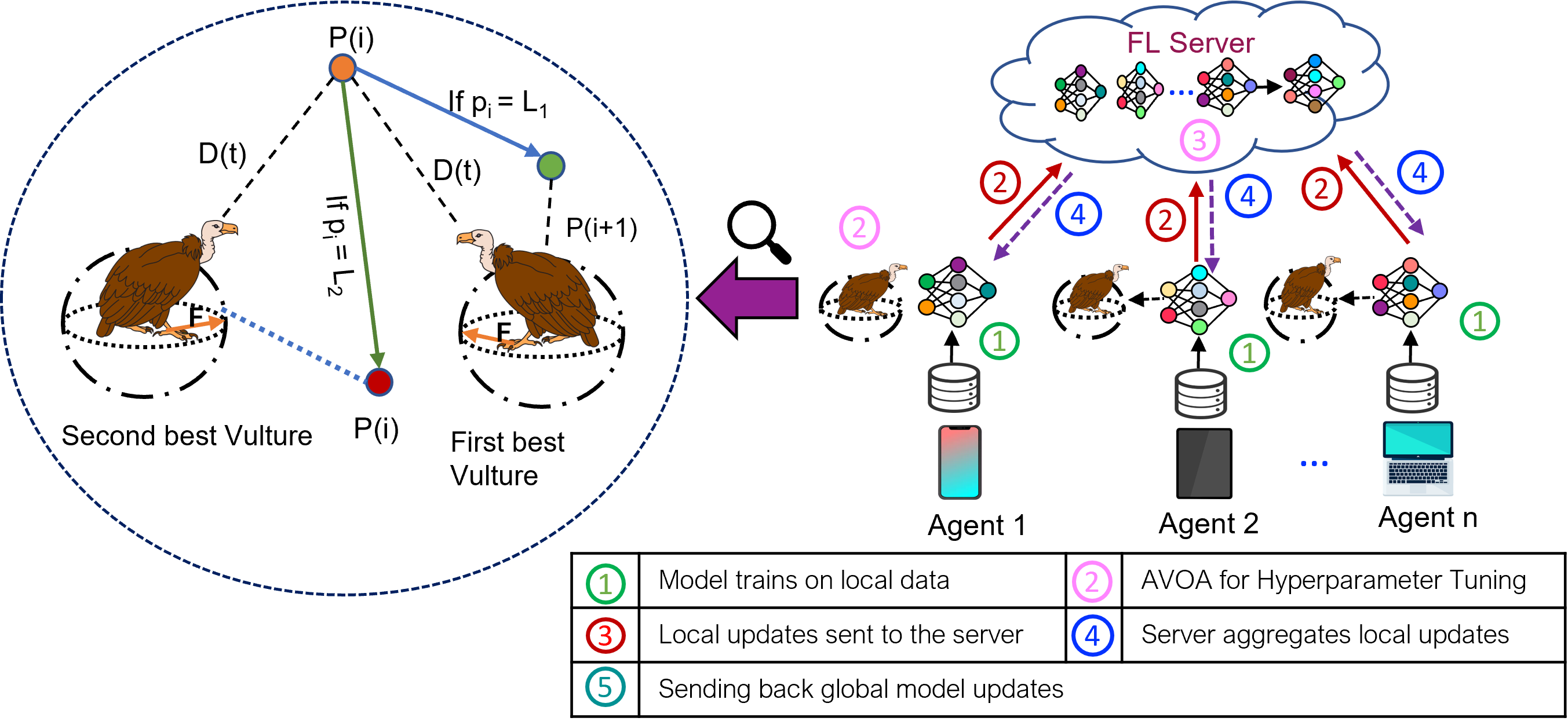}
    \caption{{\color{black}System architecture of our proposed FedAVO Algorithm.}}
    \label{fig:avo}  
\vspace{-5mm}
\end{figure*}

Unlike FedAvg \cite{mcmahan2017communication}, which employs Stochastic Gradient Descent (SGD) as its optimizer, our proposed method introduces a crucial modification aimed at automating the hyperparameter tuning process for SGD.
This adjustment has proven to yield substantial improvements in empirical performance, enabling faster convergence with a reduced number of communication rounds. To facilitate a better understanding of the approach, we have detailed the specific parameters and their notations in Table \ref{tab:notation}. Fig. \ref{fig:avo} illustrates the distinct phases of FedAVO, which consist of five key stages.\\
\textbf{Local training on client devices:} In the first phase, each client  performs independent local model training using the initial parameters received from the central server. They update their local model parameters by minimizing the loss function described in Equation \ref{eq:loss} using Stochastic Gradient Descent (SGD) on their local data. Notably, this updating process occurs in complete isolation, devoid of any communication between clients or with the server. This strict isolation ensures that clients preserve the privacy of their local data and refrain from sharing it with others.

The hyperparameter tuning phase operates concurrently with the local training phase, as illustrated in Figure \ref{fig:avo}.
To obtain hyperparameter tuning with AVO, an initial population of vultures is initialized to represent candidate solutions. Each vulture explores the problem space through random movement and evaluates its fitness based on an objective function. Here a problem space for any parameter $x$ can be denoted as $x^p={\{x_u,x_l}\}$. Where $x_l$ and $x_u$ represent the lower and upper bounds of the problem space, respectively. Vultures from the population forage through the given problem spaces for an optimal solution for a given problem. The vultures' movement is influenced by their individual experiences and interactions with other vultures within the population. Furthermore, to enhance foraging efficiency, vultures that repeatedly fail to obtain optimal solutions after a predefined number of attempts share their findings with other vultures. This iterative foraging process continues until a predetermined number of iterations or until a satisfactory solution is achieved, thereby optimizing the hyperparameters for the FL system effectively.

\begin{algorithm}
\DontPrintSemicolon
\caption{\textbf{
FedAVO (Proposed Algorithm)}
}\label{alg:2}
\textbf{On Server Side} \\
 model parameter $\omega_o$ initialized \;
\For {each training round $i=0,1,2,3 \ldots$} { 
$\quad$ $R_i\leftarrow$ random set of $q$ clients \;
$\quad$ \For{ each client $k \in \mathcal{R}_i$ } { 
$\quad$ $\quad$ $\omega_{i+1}^{k} \leftarrow$ ClientUpdate $\left(\omega_{i}, k\right)$} 
$\quad$ $\quad$$\omega_{i+1} \leftarrow \sum_{k=1}^{\mathcal{K}} \frac{n_{k}}{n} \omega_{i+1}^{k}$ \;
}
\textbf{ClientUpdate} 
The local client data $\mathcal{P}_{k}$ is split into batches of size $\mathcal{B}$ \;
Hyperparameters $\rho$ for the client are defined \;
The algorithm enters a loop for hyperparameter tuning (tuning epochs) \;
In each tuning epoch, AVO (African Vulture Optimization) is performed to optimize hyperparameters \;
A random population of vultures $Z_i$ is initialized \;
Vultures $Z_{vulture1}$ and $Z_{vulture2}$ are chosen \;
For each vulture $Z_i$, algorithm selects a region $R_i$ based on equation (\ref{eq:1}), updates fitness value $F$ using equation (\ref{eq:2}), and updates vulture's location using equations (\ref{eq:5}), (\ref{eq:6}), (\ref{eq:8}), (\ref{eq:extend}), or (\ref{eq:11}), depending on conditions and probabilities ($P_1$, $P_2$, $P_3$) \;
The best vulture's hyperparameters $\eta_i$, $\mathcal{E}_i$, $\lambda_i$, and $\beta_i$ are selected \;
A loop for local training (local epochs) is initiated, where in each local epoch, the model parameters $\omega$ are updated using SGD \;
return $\omega$ to server 
\end{algorithm}

\textbf{AVO for hyperparameter tuning:} In this study, we focus on optimizing the tuning of the following SGD hyperparameters: local epoch, which is denoted as $\mathcal{E}$, learning rate ($\eta$), momentum ($\beta$) and weight decay, which denoted as $\lambda$. The local epoch ($\mathcal{E}$), which is also an FL parameter plays an important role in convergence. Performing more local epochs on clients allows more local computation and potentially reduced communication, resulting in overall global convergence. On the contrary, a larger number of local epochs may lead each device toward the optima of its local objective as opposed to the global objective due to the heterogeneous nature of clients. This may lead to slower convergence or even cause the method to diverge and overfit. Finally, we employ objective function $E_k$ for which a problem space gets initialized with hyperparameters. The vultures forage through the problem space and search for optimal hyperparameters to solve the objective function. The optimization can be written as follows:
\begin{equation} \label{eq:loss}
   E_k(y,p) =-\sum_{c=1}^My_{o,c}\log(p_{o,c})
\end{equation}
Here, `y' represents the accurate classification for data point `o', `p' stands for the predicted probability for the same data point, and `M' represents the total number of classes.

\textbf{Server aggregation:} In this phase, the central server aggregates the locally trained model and takes a weighted average of the local updates from the selected clients. The process is repeated until global model reaches the convergence.

\vspace{-0.2cm}
\begin{table}[h]
    \centering
    \caption{Key Parameters and their Notations for FedAVO}\footnotesize
\begin{tabular}{lcl}
\hline \cellcolor{blue!40}\textbf{Parameter} & \cellcolor{blue!40}\textbf{Notation}  \\ \hline
Momentum   & $\beta_i \in [\beta,\Bar{\beta}]$ \\ 
Local epoch & $\mathcal{E}_i \in [\mathcal{E},\Bar{\mathcal{E}}] $ \\
 Learning Rate & $ \eta_i \in [\eta,\Bar{\eta}]$  \\
 Weight Decay & $ \lambda_i \in [\lambda,\Bar{\lambda}]$ \\
 Problem Space  &  $\rho$ \\
 Population size & $\alpha$  \\
\hline
\end{tabular}
    \label{tab:notation}
\end{table}
\vspace{-0.2cm}

\textbf{Global model broadcast:} Finally, the updated global model again gets sent to all selected clients and updates the local model. In the next iteration, the local data will train on the updated local model and selected hyperparameter pool by the AVO. This whole process will repeat itself until a predetermined number of communication rounds or convergence is reached.
Our Proposed FedAVO is presented with a complete pseudocode in algorithm \ref{alg:2}. 

\vspace{-0.2cm}
\section{Experimental Evaluation}
\label{sec:4}
Our study aims to improve the convergence rate and performance of FL algorithms. In order to evaluate the proposed algorithm (FedAVO), we conduct experiments and examine the convergence speed and accuracy. Furthermore, we assess if the model achieves acceptable convergence speed and accuracy after the automated HTO with AVO compared to the baseline algorithm Vanilla FedAvg \cite{mcmahan2017communication}.
In addition, we compare our experimental results with other benchmark FL algorithms such as FedProx \cite{li2020federated} and FedEnsemble \cite{zhu2021data}. 
\vspace{-0.4cm}

\subsection{Experimental Setup}
We employ a High-Performance Computing Cluster (HPCC), which has two strong GPUs (each with 2,880 extra cores) and 800 CPU cores and can process up to 34 Teraflops of computations per second (one Teraflop is one million million calculations per second). In the test code, Pytorch (stable version 2.0.0) and Keras (version 2.12.0) are used. 
For the evaluation of FedAVO, we considered four widely used image classification datasets: CIFAR-10 \cite{krizhevsky2009learning}, MNIST \cite{lecun1998gradient}, Fashion MNIST \cite{xiao2017fashion}, and LISA \cite{mogelmose2012vision}. These datasets are chosen for their frequent usage in the field of image classification. For FL training, we partition these datasets into Independent and identically distributed (IID) and Non-IID settings.
To tackle experimental datasets involving image data, we employ three distinct Convolutional Neural Network (CNN) models in our initial experiment. For MNIST, we consider a simple two-layer CNN model with 28 and 64 channels, each followed by 2 x 2 maximum pooling. This model aligns with the model used to evaluate MNIST in FedAvg \cite{mcmahan2017communication}. To ensure fairness, we match the model and parameter numbers (1,663,370 total parameters) to the FedAvg. However, for CIFAR-10 and Fashion-MNIST, we opt for three convolution layers better results. Subsequently, for training on the LISA dataset, we utilize a combination of 5 convolution layers and 3 dense layers. Since our primary focus lies in evaluating the effectiveness of our proposed hyperparameter tuning optimization method rather than achieving maximal accuracy, we find the standard CIFAR-10 model well-suited for our purposes.

\vspace{-0.05cm}
\begin{table}[htb!]
\centering
\caption{Hyperparameter Boundaries in IID AVO Problem Space}
\label{tab:4}
\footnotesize
\begin{tabular}{|c|c|c|c|}
\hline
\cellcolor{blue!40}\textbf{ID} & \cellcolor{blue!40}\textbf{Hyperparameter} & \cellcolor{blue!40}\textbf{Lower Bound} & \cellcolor{blue!40}\textbf{Upper Bound} \\
\hline
1 & Learning Rate ($\eta$) & 0.00001 & 0.01 \\
2 & Momentum ($\beta$) & 0.1 & 0.9 \\
3 & Weight Decay ($\lambda$) & 0.0001 & 0.01 \\
4 & Local epochs ($\mathcal{E}$) & 1 & 5 \\
\hline
\end{tabular}
\end{table}

\vspace{-0.4cm}

\begin{table}[htb!]
\centering
\caption{Hyperparameter Boundaries in Non-IID AVO Problem Space}\footnotesize
 \begin{tabular}{|c|c|c|c|}
\hline \cellcolor{blue!40}\textbf{ID} & \cellcolor{blue!40}\textbf{Hyperparameter} & \cellcolor{blue!40}\textbf{Lower Bound} & \cellcolor{blue!40}\textbf{Upper Bound}  \\
\hline 1 & Learning Rate ($\eta$) & 0.01 & 0.1 \\
2 & Momentum ($\beta$) & $1e^{-10}$ &$1e^{-9}$ \\
3 & Weight Decay ($\lambda$) & $1e^{-10}$  & $1e^{-8}$ \\
4 & Local epochs ($\mathcal{E}$) & 1 & 5 \\
\hline
\end{tabular}
\label{tab:5}
\end{table}




\vspace{-0.05cm}
In this study, the FedAVG method utilizes the SGD optimizer, and the hyperparameters remain unchanged from their original study. The learning rate's set to 0.01, the momentum to 0.9, and the weight decay to 0.991. On the contrary, for FedAVO, we initialize a problem space with lower bounds and upper bounds of the hyperparameters as shown in table \ref{tab:4} and table \ref{tab:5}, and we initialize the population size ($\alpha$) with 50. For both algorithms, we set the batch size to 16 and the number of clients to 10.  The local epochs number is set to 5 for each communication round in FedAvg training, but for FedAVO, we specify the local epochs number in the problem space for the HTO.
The range of hyperparameters shown in Table \ref{tab:4} and \ref{tab:5} differs based on the type of data distribution. In the case of Non-IID distribution, a higher learning rate combined with lower momentum and weight decay proves to be more effective compared to a lower learning rate with higher momentum and weight decay. In the context of FedAVO, as discussed in Section \ref{sec:method}, the limit for local epochs is maintained between 1 and 5 to prevent model overfitting. Using a significantly higher number of local epochs would lead to overfitting of the model.

\vspace{-0.25cm}
\subsection{Experimental Results}
To assess the efficacy of our FedAVO algorithm, we benchmark its performance against other state-of-the-art algorithms employing a Non-IID distribution. In our primary experiment, we divide the aforementioned datasets into $10$ clients, ensuring that each client receives 500 data points containing training images. For each of our experiments, we run upto 500 FL communication rounds.
 Furthermore, To evaluate proposed HTO, we also compare with two state-of-the-art population-based optimizers, PSO \cite{kennedy1995particle} and GWO \cite{mirjalili2014grey}. Note that, we incorporate PSO and GWO instead of AVO as an HTO
 naming them FedPSO and FedGWO. Subsequently, we conduct a performance comparison with FedAVO, demonstrating that AVO outperforms these population-based optimizers.

\begin{figure} [!htb]
\setlength{\belowcaptionskip}{-10pt}
\centering
\begin{tabular}{cccc}
\includegraphics[width=0.22\textwidth]{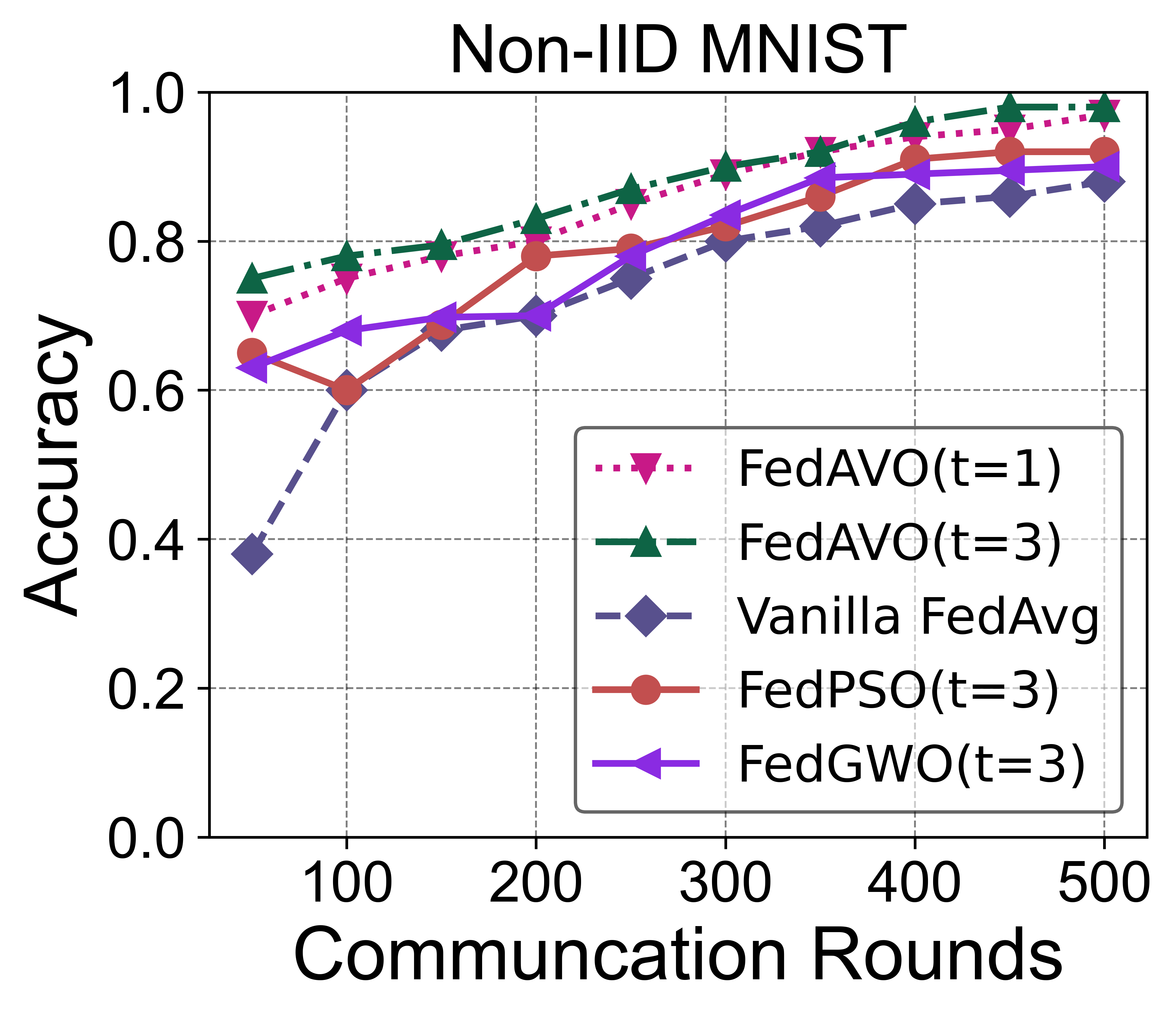}
&
\includegraphics[width=0.23\textwidth]{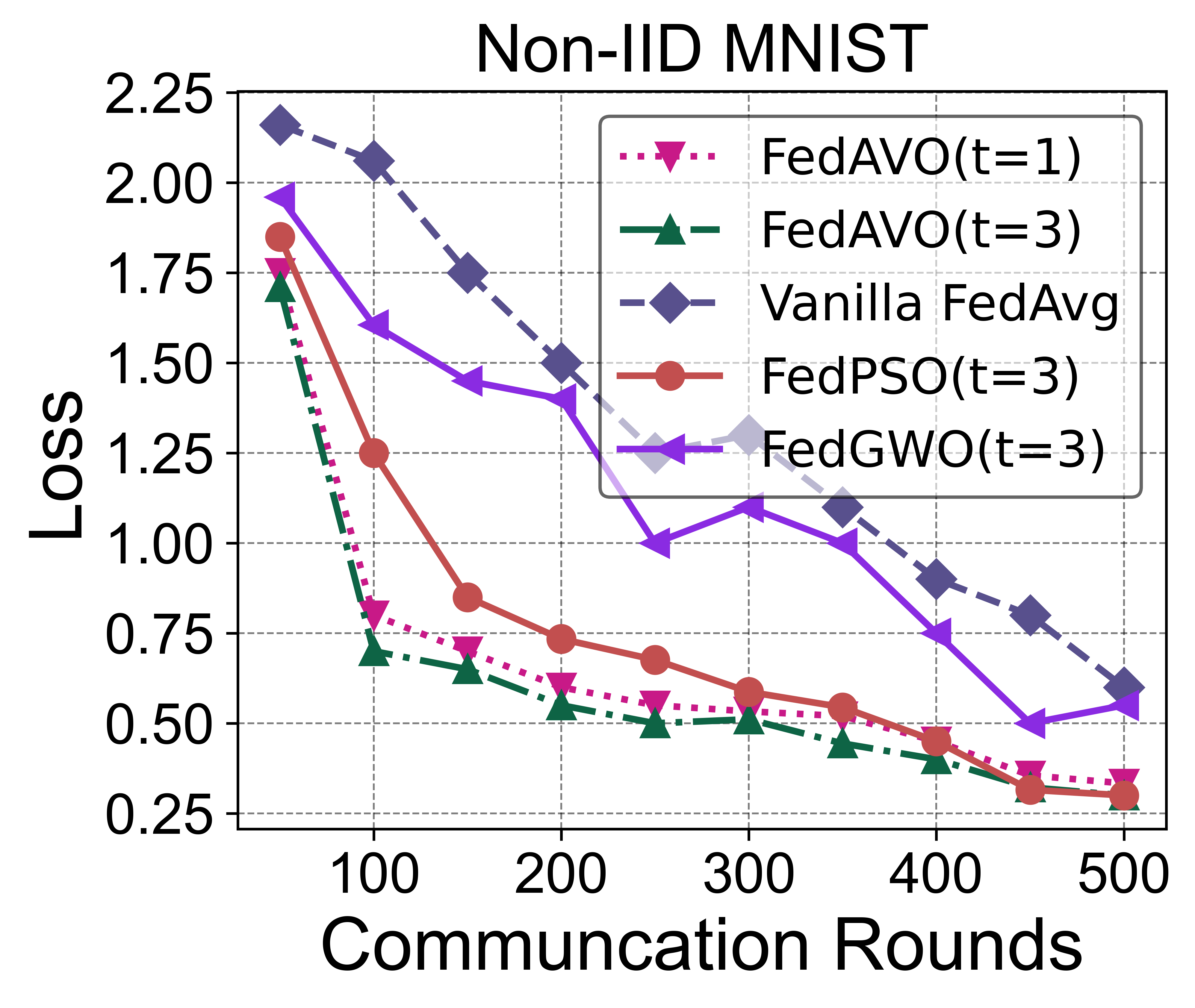}
&
\\
(a)  & (b)  \\
\end{tabular}
\caption{Comparison of model accuracy [on left] and model loss minimization [on right] considering MNIST Non-IID setting.
}
\label{fig:9}
\end{figure}


\begin{figure} [!htb]
\centering
\begin{tabular}{cc}
\includegraphics[width=0.23\textwidth]{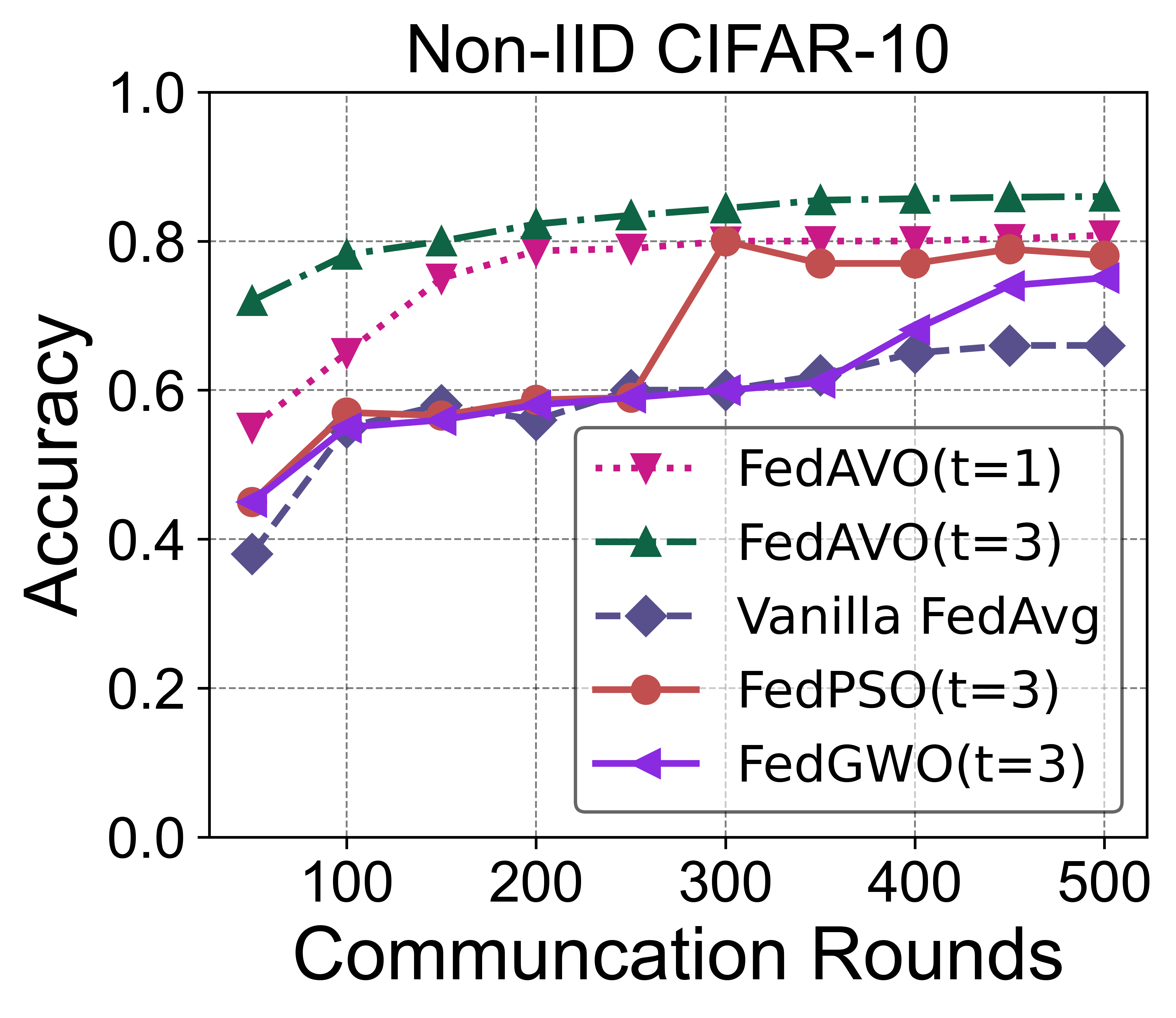}
&
\includegraphics[width=0.23\textwidth]{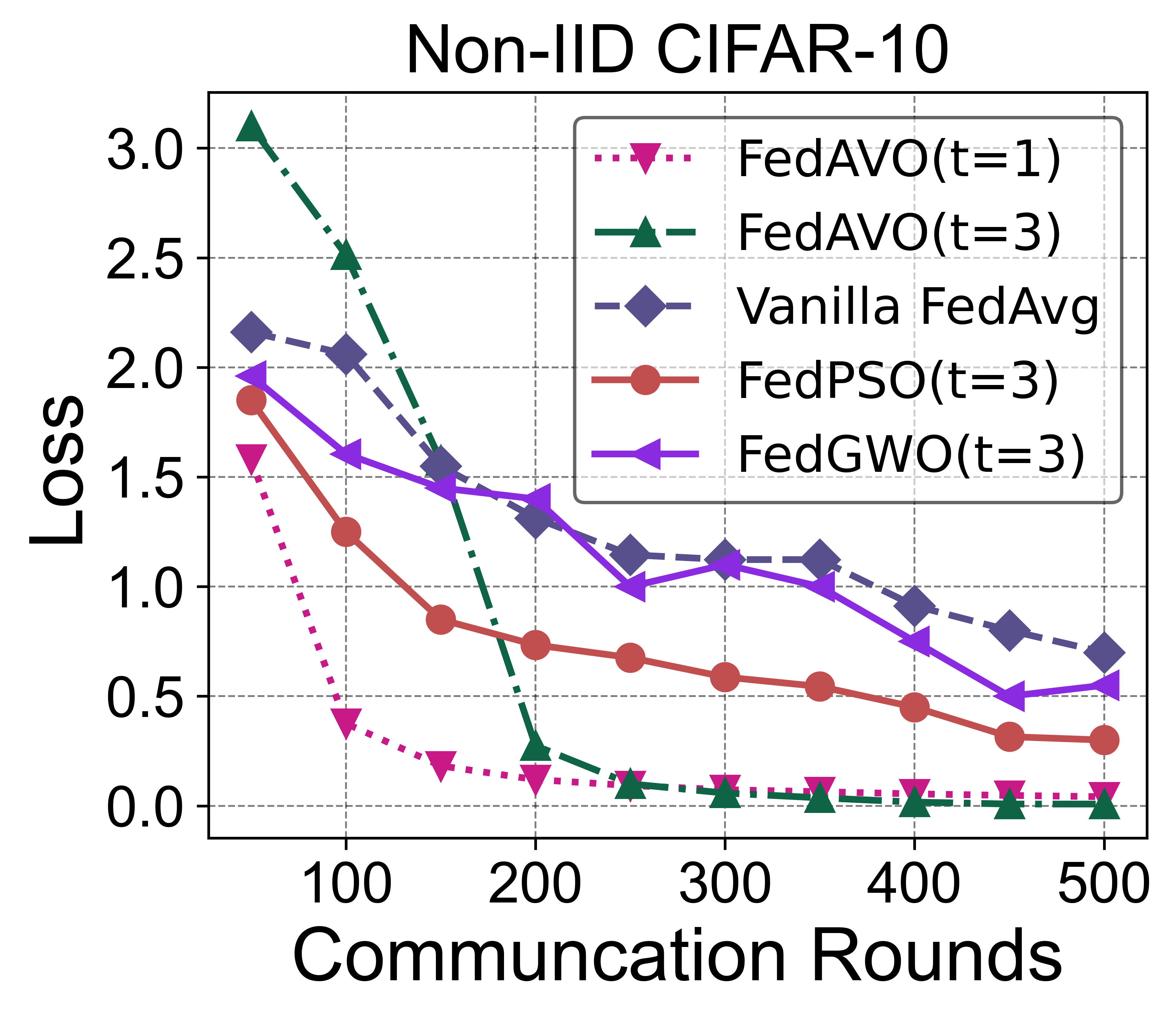}
\\
(a)  & (b)  \\
\end{tabular}
\caption{Comparison of model accuracy [on left] and model loss minimization [on right] considering CIFAR-10 Non-IID setting.
}
\label{fig:11}
\end{figure}


\begin{figure} [htb!]
\setlength{\belowcaptionskip}{-10pt}
\centering
\begin{tabular}{cc}

\includegraphics[width=0.23\textwidth]{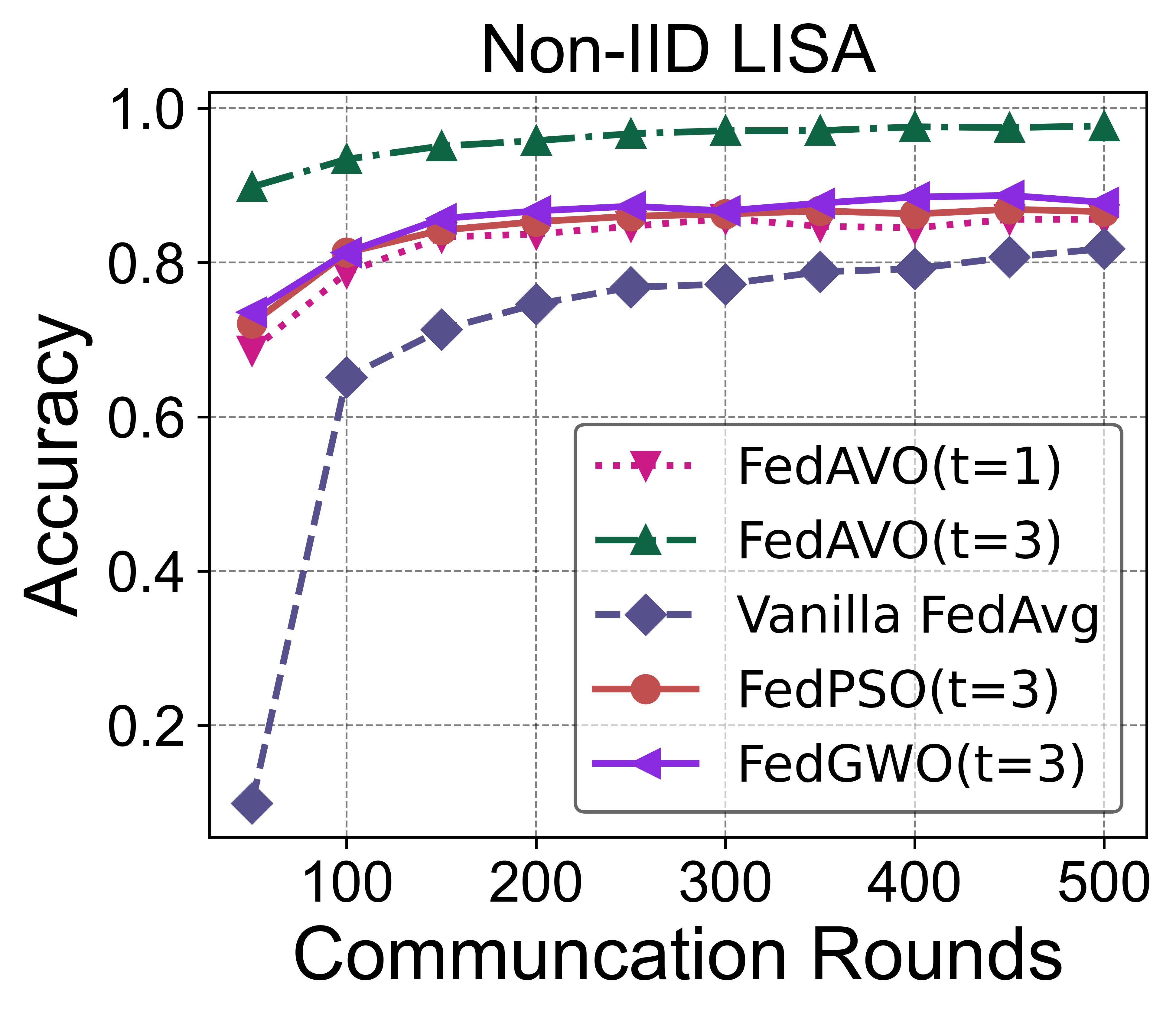}
&
\includegraphics[width=0.23\textwidth]{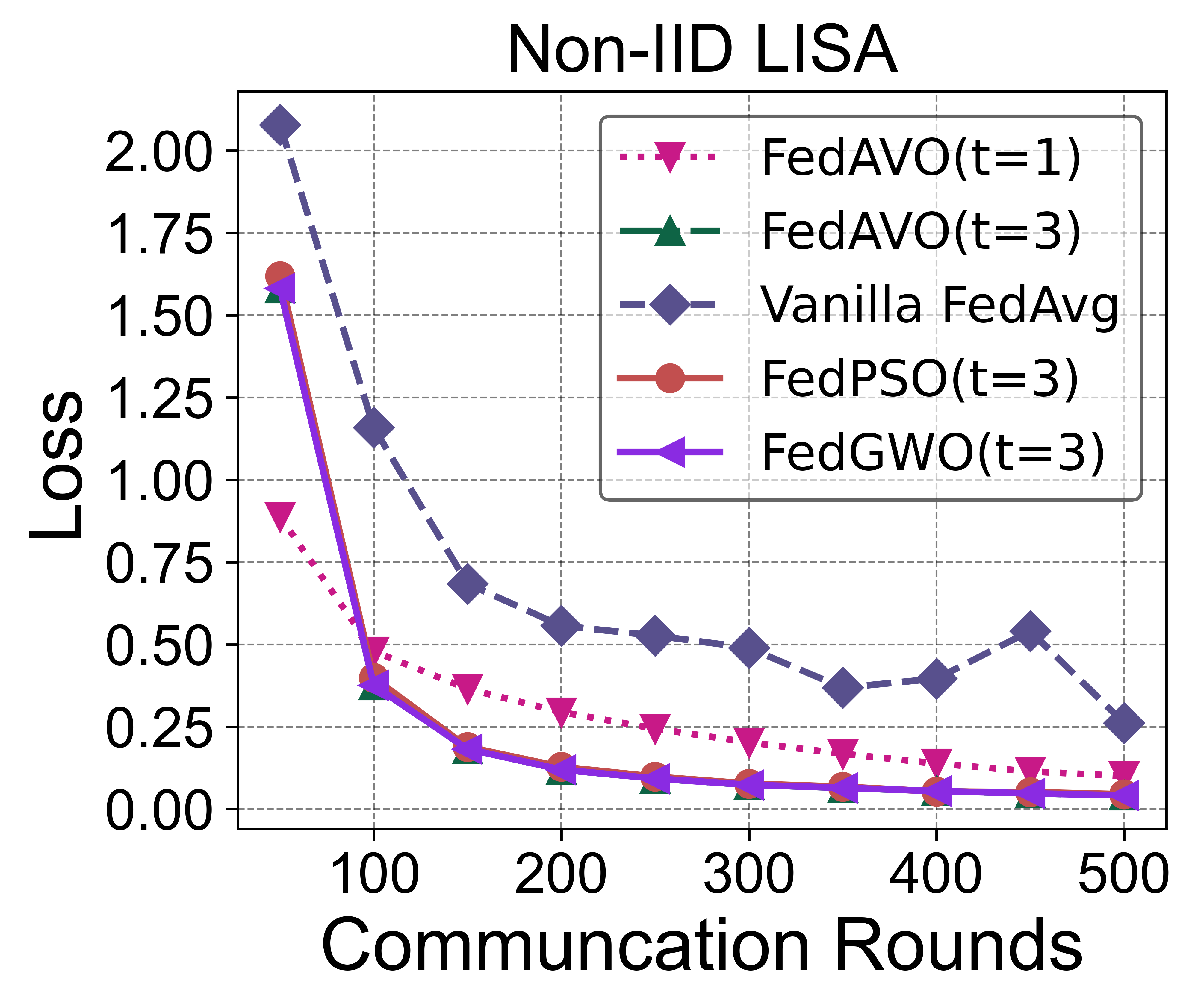 }
\\
(a)  & (b) \\
\end{tabular}
\caption{Comparison of model accuracy [on left] and model loss minimization [on right] considering LISA Non-IID setting.
}
\label{fig:fashioncomp}
\end{figure}

In Figure \ref{fig:9}, we evaluate our proposed FedAVO alongside the baseline methods Vanilla FedAvg \cite{mcmahan2017communication}, FedPSO, and FedGWO employing the MNIST dataset. From the comparison, the following inferences can be drawn: (1) Illustrated in Figure \ref{fig:9}(a) and Figure \ref{fig:9}(b), FedAVO consistently exhibits remarkable performance in terms of both global model accuracy and loss minimization in Non-IID distribution. (2) In our experiments, We explore different numbers of hyperparameter tuning epoch denoted as t to evaluate their respective performance. FedAVO outperforms other baseline algorithms by a considerable margin with tuning epoch 3 in terms of global model accuracy, Which demonstrates FedAVO performs better with a higher number of training epoch. However, even with 1 tuning epoch, our proposed method performs notably better than other FL algorithms as shown in Figure \ref{fig:9}(a). Likewise, in terms of loss minimization, FedAVO with 3 epochs shows better performance compared to Vanilla FedAVG and FedGWO shown in Figure \ref{fig:9}(b).
\begin{figure} [!htb]
\centering
\begin{tabular}{cc}
\includegraphics[width=0.23\textwidth]{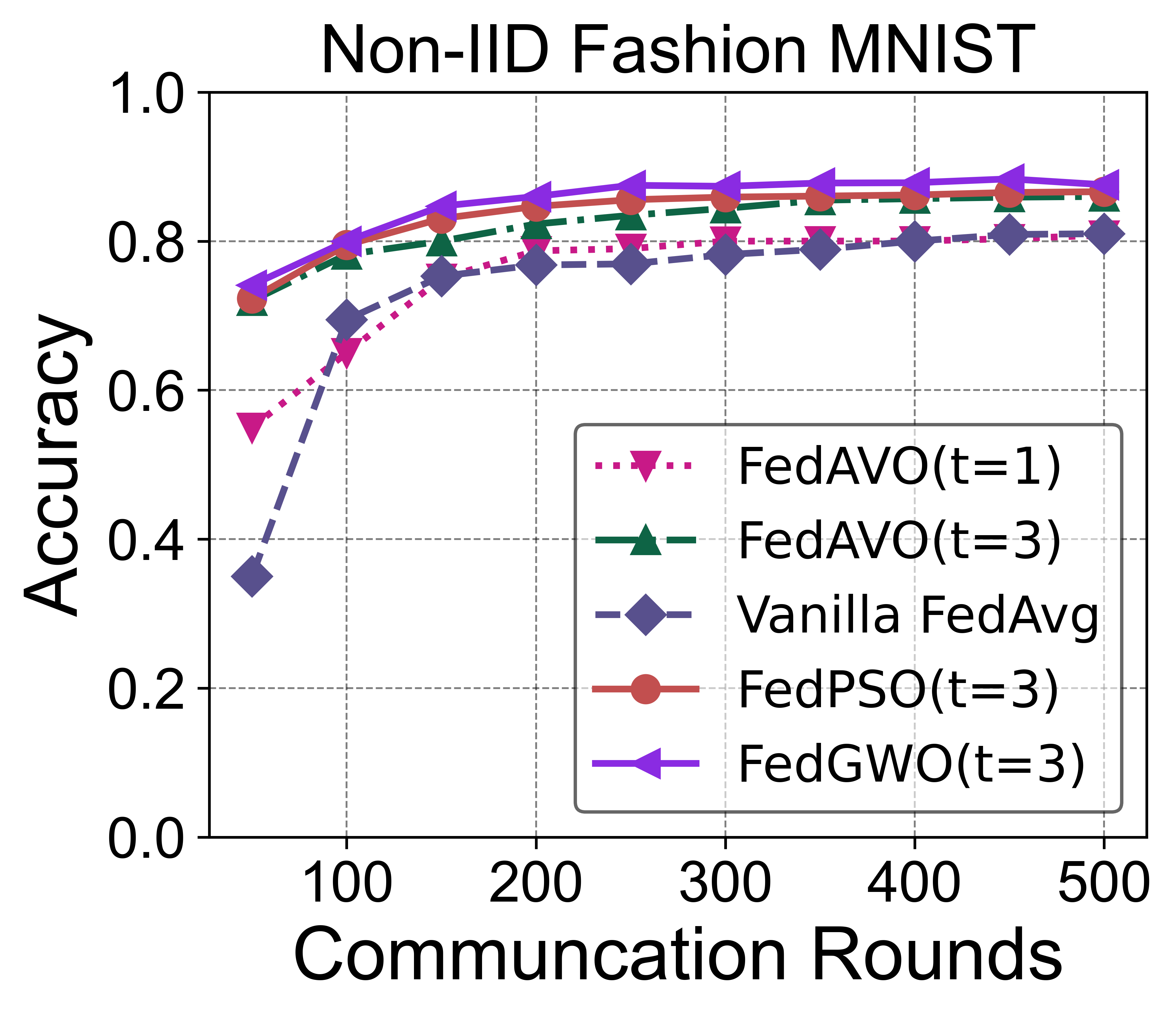}
&
\includegraphics[width=0.23\textwidth]{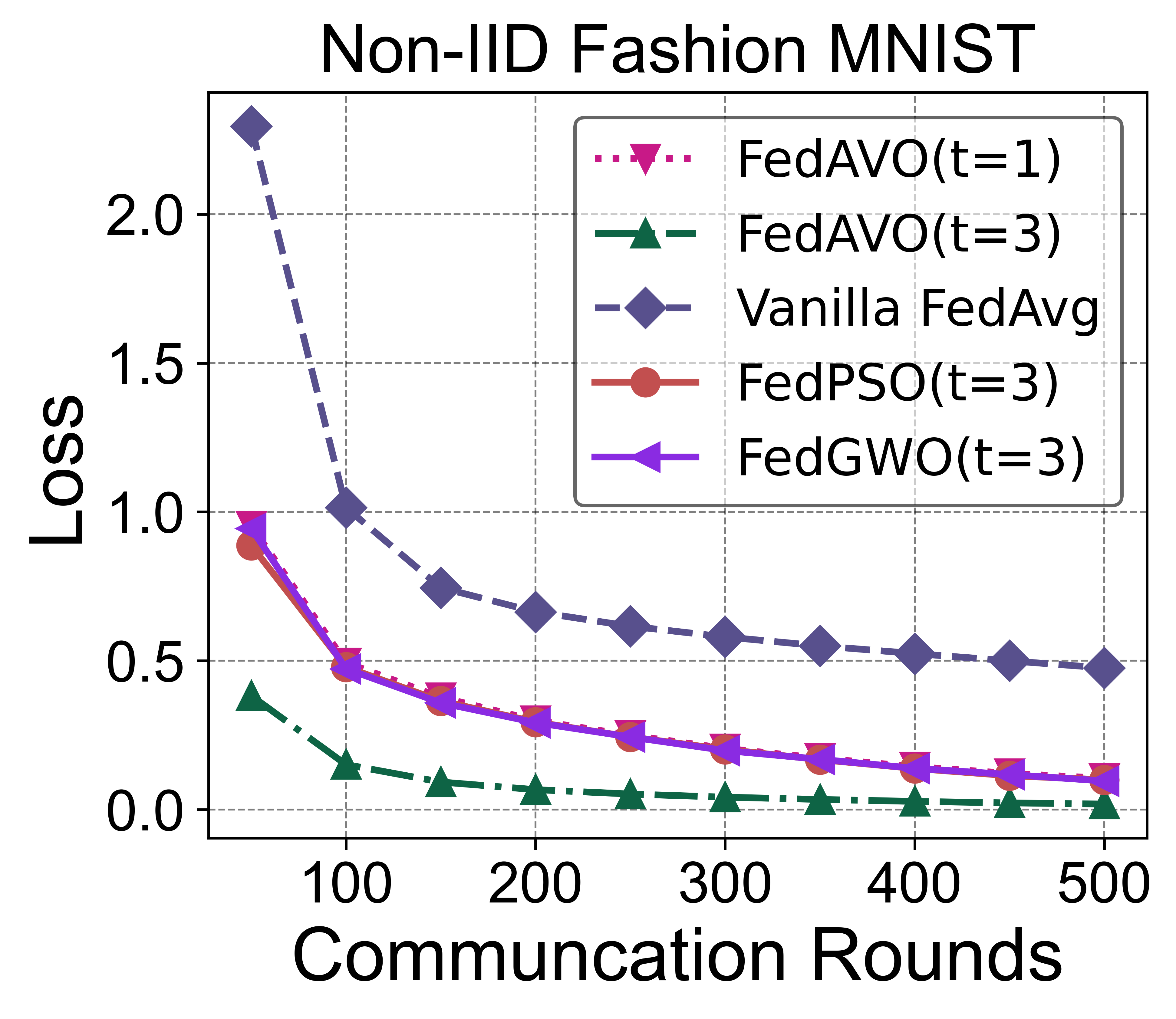}
\\
(a)  & (b)  \\
\end{tabular}
\caption{Comparison of model accuracy [on left] and model loss minimization [on right] considering Fashion MNIST Non-IID setting.
}
\label{fig:fashionmnistcor}
\end{figure}

%
Figure \ref{fig:11} depicts the performance comparison of FedAVO and other baseline FL algorithms on the CIFAR-10 dataset. From Figure \ref{fig:11}(a), we perceive that FedAVO with 3 tuning epochs performs better in terms of achieving higher global model accuracy. Due to the advantage of having both local and global search techniques, FedAVO consistently performs better than FedPSO and FedGWO. As shown in Figure \ref{fig:11}(b), Vanilla FedAvg performs poorly in terms of loss minimization. On the contrary, FedAVO minimizes loss faster than the baseline algorithms, e.g., FedPSO, FedGWO.
\begin{figure*} [htb!]
\centering
\begin{tabular}{cccc}
\includegraphics[width=0.46\textwidth]{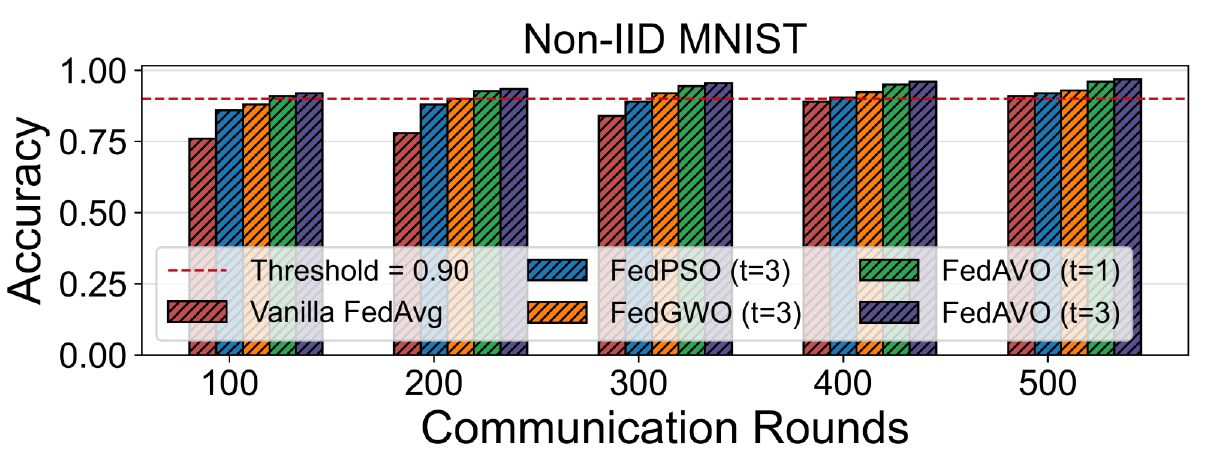}
&
\includegraphics[width=0.45\textwidth]{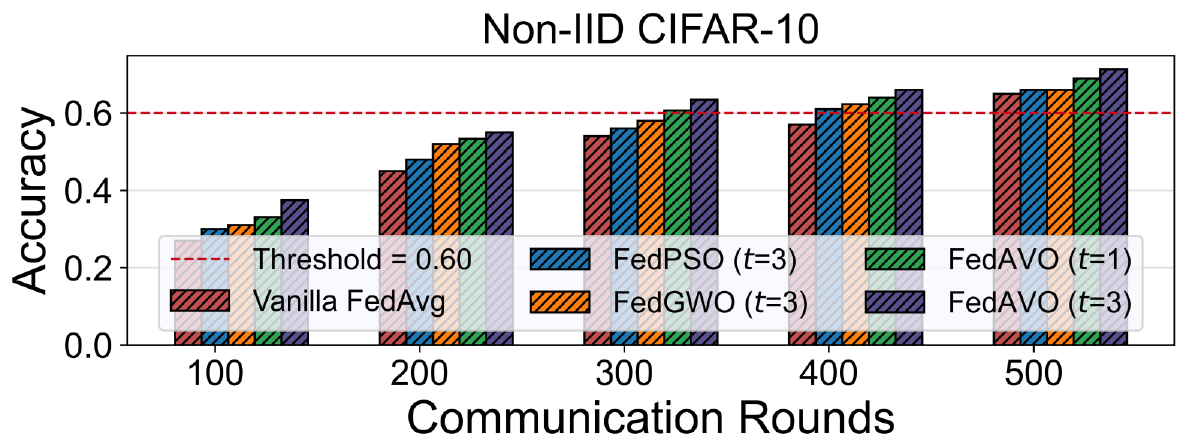}
\\
(a)  & (b) \\
\end{tabular}
\caption{Comparison of model accuracy considering (a) MNIST Non-IID and (b) CIFAR10 Non-IID setting.
 }
\label{fig:13}
\end{figure*}

\begin{table*}[htb!]
    \centering
    \footnotesize
    \caption{ Global accuracy (\%) of models trained by different FL algorithms on various datasets.}
\begin{tabular}{|c|c|c|c|c|c|c|c|c|}
\hline 
\cellcolor{blue!40}
\textbf{Dataset} & \cellcolor{blue!40}\textbf{Data Distribution} & \cellcolor{blue!40}\textbf{FEDAVG} & \cellcolor{blue!40}\textbf{FEDPROX} & \cellcolor{blue!40}\textbf{FEDENSEMBLE} & \cellcolor{blue!40}\textbf{FEDAVO (HP. Tuning 3 epochs)} \\
\hline \multirow{1}{*}{ MNIST } & IID & $97.70 \pm 2.07$ & $87.49 \pm 2.05$ & $95.85 \pm 0.68$  & $ \cellcolor{green!40}\textbf{99.67} \pm \cellcolor{green!40}\textbf{0.15}$\\
& Non-IID & $95.16 \pm 0.59$ & $90.10 \pm 0.39$ & $90.78 \pm 0.39$ & $ \cellcolor{green!40}\textbf{99.47} \pm \cellcolor{green!40}\textbf{0.25}$ \\

\hline \multirow{1}{*}{ CIFAR-10 } & IID & $67.48 \pm 0.39$ & $76.77 \pm 0.11$ & $ \cellcolor{green!40}\textbf{77.99} \pm \cellcolor{green!40}\textbf{0.23}$ & $76.34\pm 0.45$ \\
 & Non-IID & $65.13 \pm 0.25$ & $ 66.62 \pm 0.24 $ & $71.18 \pm 0.4$  & $ \cellcolor{green!40}\textbf{72.64}\pm \cellcolor{green!40}\textbf{ 0.39}$\\
\hline \multirow{1}{*}{ Fashion-MNIST} & IID & $84.37 \pm 0.9$ & $84.77 \pm 0.31$ & $83.89 \pm 0.33$ & $ \cellcolor{green!40}\textbf{89.34}\pm \cellcolor{green!40}\textbf{0.35}$ \\
 & Non-IID & $82.00 \pm 0.43$ & $ 83.62 \pm 0.77 $ & $85.28 \pm 0.35$  & $ \cellcolor{green!40}\textbf{87.01}\pm \cellcolor{green!40}\textbf{ 0.5}$\\
 \hline \multirow{1}{*}{ LISA } & IID & $92.48 \pm 0.39$ & $94.24 \pm 0.5$ & $93.43 \pm 0.25$ & $ \textbf{96.34}\pm \cellcolor{green!40}\textbf{0.4}$ \\
 & Non-IID & $88.34 \pm 0.34$ & $ 89.5 \pm 0.45 $ & $92.88 \pm 0.4$  & $\textbf{94.00}\pm \cellcolor{green!40}\textbf{ 0.45}$\\
 \hline
\end{tabular}
    \label{tab:accuracy}
\end{table*}

We also conduct empirical analysis on the LISA non-IID dataset, depicted in Figure \ref{fig:fashioncomp}(a) and \ref{fig:fashioncomp}(b). In this scenario, FedAVO and the baseline algorithms demonstrate comparable performance in terms of minimizing the loss function. However, regarding global model accuracy, FedAVO outperformed the other baseline algorithms, achieving superior results with only three hyperparameter tuning epochs.
Finally, Figures \ref{fig:fashionmnistcor}(a) and \ref{fig:fashionmnistcor}(b) illustrate the performance of FedAVO on the Fashion MNIST dataset in Non-IID distribution. Notably, Both FedPSO and FedGWO show performance par with FedAVO with 3 tuning epoch. However, FedAVO shows marginal improvement in terms of loss minimization illustrated in \ref{fig:fashionmnistcor}(b).

In Figure \ref{fig:13}, we set a 90\% accuracy threshold to compare FedAVO against state-of-the-art FL methods, measuring the accuracy gain over multiple communication rounds. Algorithms that surpass the threshold in fewer communication rounds are deemed to have a higher convergence rate. For the MNIST dataset, under Non-IID distribution FedAVO outperforms FedAvg by reducing communication rounds by 80. Remarkably, FedAVO achieves the specified threshold within 100 communication rounds. Furthermore, on the CIFAR-10 dataset, as depicted in Figure \ref{fig:13}(b), FedAVO demonstrates expedited convergence. To further evaluate FedAVO, we set a 60\% threshold for CIFAR-10. In contrast to FedAvg, FedAVO exhibits robust performance in CIFAR-10 Non-IID scenarios, reaching the threshold within 300 communication rounds compared to FedAvg's 450 rounds. For the CIFAR-10 dataset, FedAVO effectively reduces communication rounds by up to 150. Based on our extensive result analysis, we can conclude that the implementation of FedAVO within FL techniques can halve or even more the communication round requirements, showcasing its efficacy in enhancing the convergence rate.

We compare our proposed FedAVO's performance in terms of global accuracy with other state-of-the-art FL algorithms and summarize the empirical analysis in Table \ref{tab:accuracy}. Additionally, we showcase our empirical analysis in IID settings to provide a comprehensive overview of FedAVO's performance. For CIFAR-10 dataset, FedAVO increases the performance gain compared to FedAvg from 67.48\% to 76.34\% on IID data and 65.13\% to 72.64\% on non-IID data. With MNIST dataset FedAVO shows a 6\% increase in global accuracy compared to FedAvg. Although with CIFAR-10 IID distribution, FEDENSEMBLE shows slightly better performance by leveraging the benefit of having ensemble predictions from the user models, FedAVO shows substantially better performance on the Non-IID distribution of CIFAR-10. From the mentioned table, we can also observe that for MNIST and CIFAR-10 Non-IID distribution, FedAVO achieves 99.47\% and 72.64\% accuracy, respectively.
Our experimentation on the Fashion-MNIST and LISA datasets offers additional evidence supporting our claim that FedAVO consistently yields enhanced performance in terms of global accuracy. In heterogeneous systems for both of the aforementioned datasets, FedAVO shows a 5.5\% increase on average in terms of global accuracy.

\section{Conclusion}
\label{sec:5}
This research focuses on reducing communication overhead in federated learning (FL) by fine-tuning the hyperparameters of consumer electronics devices involved in edge learning. FedAVO, our advanced FL algorithm, optimizes hyperparameter adjustments to significantly enhance FL performance. We conducted a thorough assessment on benchmark datasets, comparing FedAVO against state-of-the-art FL algorithms. The results unequivocally establish FedAVO's superiority over widely adopted FL algorithms like FedAvg, FedPSO, FedEnsemble, and FedGWO, showcasing an average global accuracy improvement of up to 6\%. Notably, FedAVO exhibits exceptional capabilities in reducing communication rounds, achieving an average reduction of 30 rounds for the MNIST dataset compared to FedAvg and a substantial reduction of around 150 rounds for the CIFAR-10 dataset. Even when compared to FedGWO and FedPSO, FedAVO significantly reduces communication rounds by up to 50 for CIFAR-10 and an average reduction of 20 for MNIST. Moreover, FedAVO's adaptability extends to various FL contexts and time-sensitive systems, promising performance enhancements and reduced round requirements in practical real-world scenarios. Future research will explore further improvements in network communication performance by addressing challenges like the risk of local minima and facilitating P2P-AVO communication among clients, potentially through the implementation of dynamic multi-vulture AVO.




\bibliographystyle{ieeetr}

\bibliography{Main}
\end{document}